*Article*

# Pareto Optimization or Cascaded Weighted Sum: A Comparison of Concepts

## Wilfried Jakob [1],* and Christian Blume [2]

[1] Karlsruhe Institute of Technology (KIT), Institute of Applied Computer Science (IAI), P.O. Box 3640, Karlsruhe 76021, Germany

[2] Cologne University of Applied Sciences, Institute of Automation and Industrial IT, Steinmüllerallee 1, Gummersbach 51643, Germany; E-Mail: blume@gm.fh-koeln.de

* Author to whom correspondence should be addressed; E-Mail: wilfried.jakob@kit.edu; Tel.: +49-721-608-24663; Fax: +49-721-608-22602.

**Abstract:** Looking at articles or conference papers published since the turn of the century, Pareto optimization is the dominating assessment method for multi-objective nonlinear optimization problems. However, is it always the method of choice for real-world applications, where either more than four objectives have to be considered, or the same type of task is repeated again and again with only minor modifications, in an automated optimization or planning process? This paper presents a classification of application scenarios and compares the Pareto approach with an extended version of the weighted sum, called *cascaded weighted sum*, for the different scenarios. Its range of application within the field of multi-objective optimization is discussed as well as its strengths and weaknesses.

## 1. Introduction

Most nonlinear real-world optimization problems require the optimization of several objectives and usually at least some of them are contradictory. A simple example of two conflicting criteria is the *payload* and the *traveling distance* with a given amount of fuel, which cannot be maximized both at the same time. The typical solution of such a problem is a compromise. A good compromise is one where one of the criteria can be improved only by worsening at least one of the others. This approach is called *Pareto optimization* [1], and the set of all good compromises is called *Pareto optimal solutions* or *non-dominated solutions*. In practice, usually only one solution is required. Thus, multi-objective optimization based on Pareto optimality is divided into two phases: At first, the set of Pareto optimal solutions is determined, out of which one must be chosen as the final result by a human decision maker according to more or less subjective preferences. This is in contrast to single-objective optimization tasks, where no second selection step is required.

Most population-based search procedures, like evolutionary algorithms, particle swarm or ant colony optimization, require a single quality value called e.g., fitness in the context of evolutionary algorithms. This may be one reason for the frequent aggregation of different optimization criteria to a single quality value. Two methods, the frequently used *weighted sum* and the *ε-constrained method*, are described briefly. Another commonly used method is to express everything in costs. On closer inspection, it becomes apparent that this is equal to the weighted sum approach using suitable weights. Additionally, the conversion into costs requires an artificial redefinition of the original goals and this is often not really appropriate. Thus, most multi-objective optimization problems have meanwhile been solved based on Pareto optimization, at least in academia.

The computational effort to determine all or at least most of the Pareto front increases significantly with the number of conflicting objectives, as will be shown later in the paper. However, what if the complete Pareto front is not needed at all, because the area of interest is already known? In this paper we will introduce an aggregation method called the *cascaded weighted sum* (CWS) and discuss application scenarios, where aggregation methods like the CWS can compete with Pareto-optimality-based approaches. Not to be misunderstood: We agree that in many fields of application, Pareto optimization is the appropriate method for multi-objective problems. Although we will concentrate on evolutionary multi-objective optimization later in the paper, the issues discussed here can be applied to other global optimization procedures as well and especially to those, which optimize a set of solutions simultaneously instead of just one.

The paper is organized as follows: In Section 2 the basics of Pareto optimization are described, followed by the weighted sum and the ε-constrained method, including a brief discussion of their properties. In Section 3, the cascaded weighted sum is introduced. Section 4 starts with a classification of application scenarios, gives some examples, and discusses the question for which scenario which method is suited better or how they can complement each other. The paper closes in Section 5 with a summary and a conclusion.

## 2. Short Introduction to Pareto Optimization and Two Aggregation Methods

Based on Hoffmeister and Bäck [2], and the notation of Branke *et al.* [3], a multi-objective optimization problem is the task of maximizing a set of $k$ ($>1$) usually conflicting *objective functions* $f_i$ simultaneously, denoted by *maximize {...}*:

$$\text{maximize } \{f_1(x), f_2(x), \ldots, f_k(x)\},\ x \in S$$
$$f_i : S \subseteq S_1 \times \ldots \times S_n \to \Re,\ S \neq \varnothing \tag{1}$$

The focus on maximization is without loss of generality, because $\min\{f(x)\} = -\max\{-f(x)\}$. The nonempty set $S$ is called the *feasible region* and a member of it is called a *decision (variable) vector* $x = (x_1, x_2, \ldots, x_n)^T$. As it is of no further interest here, we do not describe the constraints forming $S$ in more detail. Frequently, the $S_i$ are the set of real or whole numbers or a subset thereof, but they can be any arbitrary set as well. *Objective vectors* are images of decision vectors, consisting of *objective (function) values* $z = f(x) = (f_1(x), f_2(x) \ldots, f_k(x))^T$. Accordingly, the image of the feasible region in the objective space is called the *feasible objective region* $Z = f(S)$. Figure 1 illustrates this.

**Figure 1.** Feasible region $S$ and its image, the feasible objective region $Z$ for $n = k = 2$. The set of weakly Pareto optimal solutions is shown as a bold green line in the diagram on the right. The subset of Pareto optimal solutions is the part of the green line between the black circles. The ideal objective vector $z^*$ consists of the upper bounds of the Pareto set.

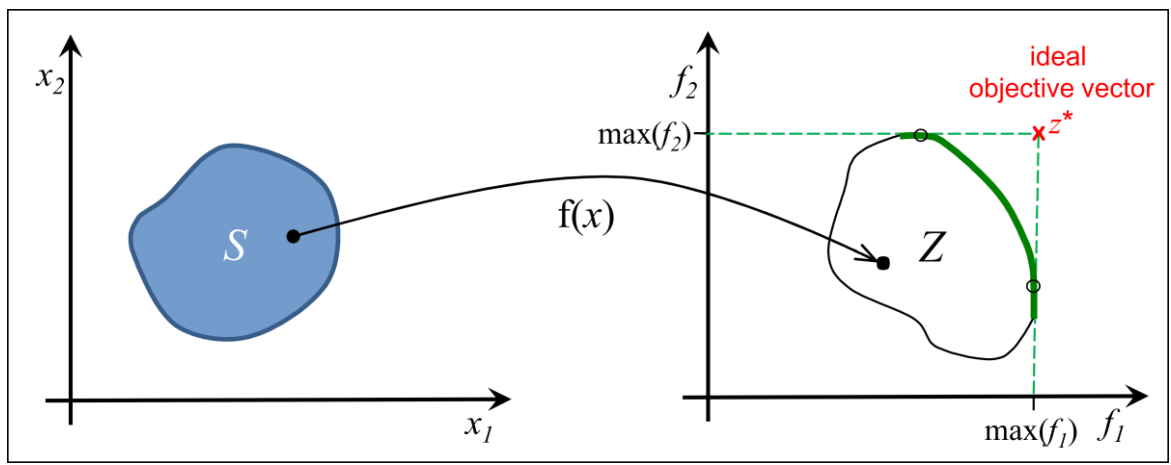

In the following sections Pareto optimization and two frequently used aggregation methods, which turn a multi-objective problem into a single-objective task, are introduced and compared in the end.

## 2.1. Pareto Optimization

A decision vector $x \in S$ dominates another vector $y \in S$, if

$$\forall i \in \{1,2,\ldots,k\}: f_i(x) \geq f_i(y) \quad and$$
$$\exists j \in \{1,2,\ldots,k\}: f_j(x) > f_j(y)$$

(2)

A decision vector $x' \in S$, which is not dominated by any other $x \in S$, is called *Pareto optimal*. The objective vector $z' = f(x')$ is Pareto optimal, if the corresponding decision vector is Pareto optimal and the corresponding sets can be denoted by $P(S)$ and $P(Z)$. The set of weakly Pareto optimal solutions, which is a superset of the set of Pareto optimal solutions, is formed by decision vectors, for which the following applies: An $x' \in S$ is called *weakly Pareto optimal*, if no other $x \in S$ exists such that $f_i(x) > f_i(x')$ for all $i = 1, \ldots, k$. As the set of Pareto optimal solutions consists of decision vectors only, which are not dominated, they can be regarded as the set of good compromises mentioned in the introduction. It follows from the definition that they are located on the border of the feasible objective region, as shown in the right part of Figure 1. The figure also illustrates the concept of weakly Pareto optimal solutions lying on the part of the green line outside of the section bounded by the black circles in the given example. It should be stated that the set of Pareto optimal solutions does not need to be as nicely shaped as shown in Figure 1; it may also be non-convex and disconnected.

The upper bounds of the Pareto optimal set can be obtained by maximizing the $f_i$ individually with respect to the feasible region. This results in the *ideal objective vector* $z* \in \Re^k$, an example of which is shown for the two-dimensional case in the right part of Figure 1. The lower bounds are usually hard to determine, see [3]. Although Pareto-based search methods can provide valuable estimations of the ranges of the objectives for practical applications, they are not suited for an exact determination of their lower and upper bounds.

According to [4], constraints in the objective space are handled as follows: A solution $x$ *constrained-dominates* a solution $y$, if any of the three conditions is satisfied:

- Solution $x$ is feasible and $y$ is not.
- Both solutions are feasible and $x$ dominates $y$.
- Both solutions are infeasible, but $x$ has a smaller constrained violation than $y$. If more than one constraint is violated, the violations are normalized, summed up, and compared.

Hereinafter, the term *Pareto optimization* is used for an optimization procedure employing Pareto optimality to assess and compare generated solutions.

## 2.2. Weighted Sum

One of the probably most often used assessment methods besides Pareto optimality is the weighted sum, which aggregates the objective values to a single quality measure. As the objective functions frequently have different scales, they are usually normalized. This can be done for example by using Equations (3) or (4) when minimizing and maximizing the objectives, respectively:

$$f_i^{norm} = \frac{\max(f_i) - f_i}{\max(f_i) - \min(f_i)} \text{ for objectives to be minimized} \tag{3}$$

$$f_i^{norm} = 1 - \frac{\max(f_i) - f_i}{\max(f_i) - \min(f_i)} \text{ for objectives to be maximized} \tag{4}$$

The bounds of the objective function $f_i$ can be estimated or are the result of a maximization of each function individually in case of $\max(f_i)$. For the calculation of the weighted sum as shown in Equation (5), a weight $w_i$ has to be chosen for every objective:

$$\text{maximize} \sum_{i=1}^{k} w_i f_i^{norm}(x), \; x \in S \; \text{ where } \; w_i > 0 \text{ for all } i = 1, \dots, k \text{ and } \sum_{i=1}^{k} w_i = 1 \tag{5}$$

By varying the weights, any point of a convex Pareto front can be obtained. Figure 2 illustrates this: The straight line corresponding to the chosen weights $w_1$ and $w_2$ is moved towards the border of the feasible objective region during the optimization process and becomes a tangent in point P. The solutions found are Pareto optimal, see [5].

On the other hand, it is possible that parts of the Pareto front cannot be found in case of a non-convex problem. This is illustrated in Figure 3: the part between points A and B of the Pareto front cannot be obtained for any weights. This is a serious drawback.

**Figure 2.** By using appropriate weights, every point of a convex Pareto front can be achieved by the weighted sum. Here, point **P** can be obtained for the weights $w_1$ and $w_2$. The arrows show the movement direction of points where the largest quality gain is obtained.

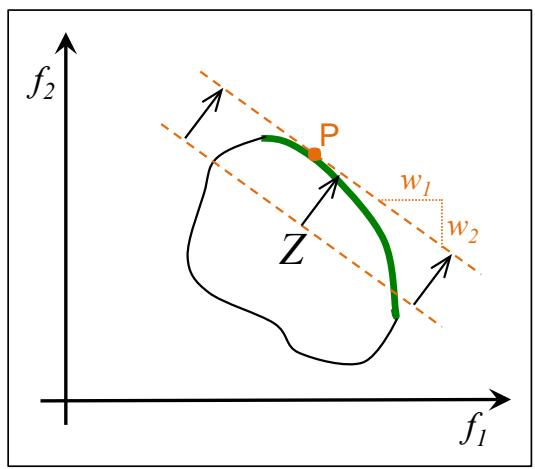

**Figure 3.** For non-convex Pareto fronts, it is possible that parts of the front can not be obtained by the weighted sum. The region between points **A** and **B** is an example of this serious draw back of this aggregation method.

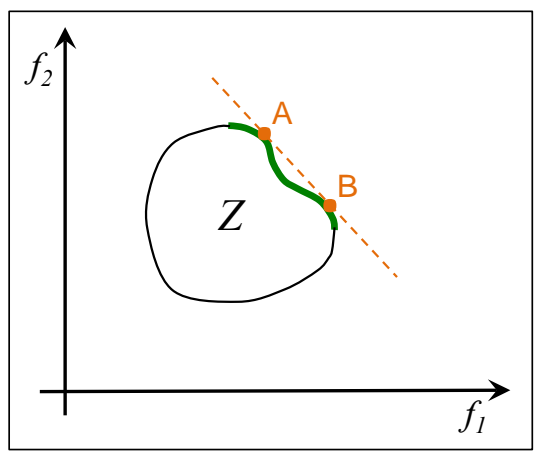

As mentioned above, the weighted sum is often used for practical applications. Reasons are the simplicity of its application and the easy way to integrate restrictions, which are beyond pure limitations of the feasible region. Examples are scheduling tasks, where the jobs to be scheduled have due dates for finalization. Thus, delays can occur and it is not sufficient to tell the search procedure that this constrained violation is an infeasible solution by e.g., rejecting it. Instead, the search must be guided out of the infeasible region by rewarding a reduction of the violation. In the example given, this can be done by counting the number of jobs involved and summing up the amounts of delays, see e.g., [6]. These two key figures can either become new objectives or can be treated as penalty functions. As they do not represent wanted properties and as a low number of objectives is preferable, penalty functions are the method of choice. They can be designed to yield values between zero (maximal violation) and one (no violation). The results of all penalty functions serve as factors by which the weighted sum is multiplied. As a result, the pure weighted sum turns into a raw quality measure, which represents the solution quality of the problem without constraints, while the final product represents the solution for the task with its constraints. Figure 4 shows a typical example of such a penalty function. If the maximum value of constraint violation is hard to predict, as it is often the case, an exponential function can be chosen. A value of delay *dp* for poor solutions can usually be estimated roughly and the exponential function is attributed such that it yields a value of ⅓ in this case.

**Figure 4.** Example of a penalty function. It turns constraint violations into a penalty value between 1 and 0, which serves as a factor for decreasing the weighted sum.

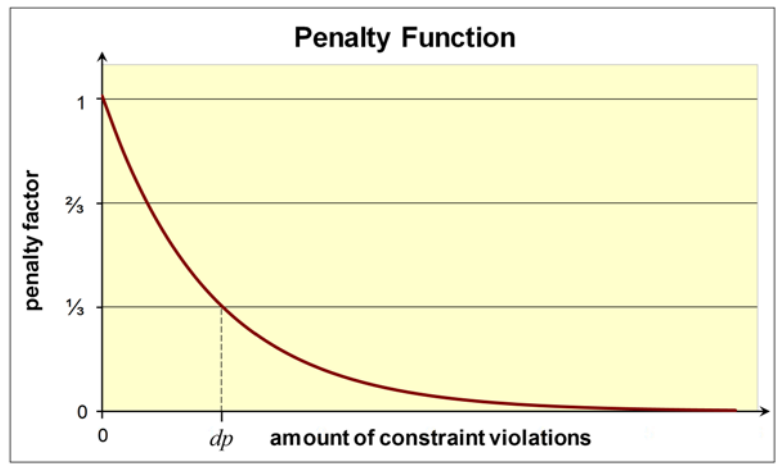

### 2.3. ε-Constrained Method

The ε-constrained method is based on the optimization of one selected objective function $f_j$ and treating the others as constraints [7]. The optimization problem now has the form

$$\text{maximize } f_j(x), \quad x \in S, \quad j \in \{1,...,k\}$$
$$f_i(x) \geq \varepsilon_i \text{ for all } i = 1, ..., k, \ i \neq j \tag{6}$$

The $\varepsilon_j$ are the lower bounds for those objective functions that are treated as constraints. For practical applications, the appropriate bounds must be selected carefully. In particular, it must be ensured that the bounds are within the feasible objective region, because otherwise the resulting problem would have no solutions. Osyczka gives suggestions for a systematic selection of values for the $\varepsilon_j$ and illustrates them with sample applications [8].

Figure 5 gives an example based on the feasible region shown in Figure 3. In Figure 5 $f_2$ is treated as a constraint with the lower bound $\varepsilon_2$. Thus, the remaining Pareto front is the section between the points F1 and F2. The figure also shows the main movement direction of solutions in *Z* that have exceeded the threshold $\varepsilon_2$. The main movement direction results from the optimization. Move components up- and downwards are also possible, but are not considered by the assessment procedure of this method as long as they do not drop below $\varepsilon_2$. A too large value of the constraint like $\varepsilon_{bad}$ would make the problem unsolvable.

A decision vector $x' \in S$ is Pareto-optimal, if and only if it solves Equation (6) for every $j = 1,...,k$, where $\varepsilon_i = f_i(x')$ for $i = 1,...,k, \ i \neq j$, see [9]. This means that *k* different problems must be solved for every member of the Pareto front, which can be expected to be computationally costly. If the task can be relaxed to weak Pareto optimality, only one solution of Equation (6) per member of the front is required [9]. On the other hand, the method does not require convexity for finding any Pareto-optimal solution.

> **Figure 5.** Restricted objective region using the $\varepsilon$-constrained method. The hatched region is excluded due to the lower bound $\varepsilon_2$. The remaining Pareto front is limited by **F1** and **F2**. For too large bounds like $\varepsilon_{bad}$, the problem becomes unsolvable.

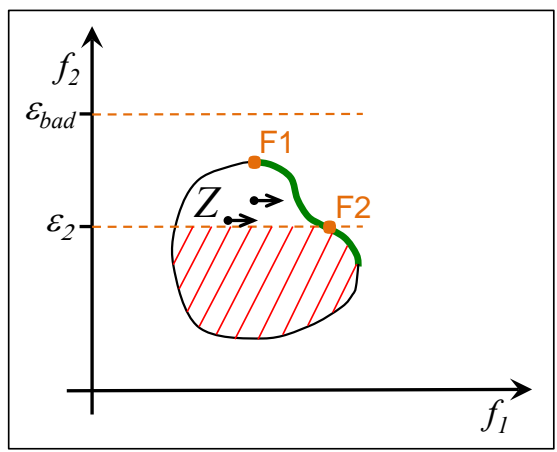

## 2.4. Summary

The two aggregation procedures (More aggregation methods can be found in [9].) can both find any Pareto-optimal solution for convex problems and the $\varepsilon$-constrained method can do that for non-convex tasks, too. This advantage of the $\varepsilon$-constrained method goes at the expense of higher computational costs: For finding a Pareto-optimal solution, the $\varepsilon$-constrained method needs to solve $k$ different problems, whereas the weighted sum needs to solve just one per element of the Pareto set.

Another important issue is the manageability. Depending on the problem, it can be assumed that it is easier for experts in the application area to estimate lower bounds for objectives than weights, which are more abstract. Experts are usually familiar with objective values and they are intelligible for them as such.

For optimization procedures working with a set of solutions, frequently called population, like Evolutionary Algorithms (EAs), Ant Colony (ACO) or Particle Swarm Optimization (PSO), or the like, the advantage for multi-objective optimization is the ability to determine, in principle, the entire Pareto front at once. Of course, an optimization procedure must be adapted such that it spreads the population to the Pareto front as best as possible instead of concentrating on some areas of it. Examples of such adapted procedures in the EA field are the *non-dominated sorting Genetic Algorithm* (NSGA-II) [10], the *Strength Pareto EA* (SPEA-2) [11], or the *S-metric selection evolutionary multi-objective optimization algorithm* (SMS-EMOA) [12], to mention only a few. These combinations of population-based optimization procedures and Pareto optimality in general will estimate the Pareto front roughly at the minimum, with less computational effort than it can be done using the aggregation methods introduced for the assessment of the individuals of the same population-based methods. In the latter case, about as many runs would be required as solutions should occupy the Pareto front. Thus, these algorithms specialized for finding as much of the Pareto front as possible in one run are the methods of choice for new problems, where little is known in advance and where a human decision maker is available to make the final choice.

## 3. Cascaded Weighted Sum

We have been using this extended version of the weighted sum since the inception of our Evolutionary Algorithm GLEAM (General Learning and Evolutionary Algorithm and Method) in 1990 [13], as we considered this assessment method a convenient way to steer the evolutionary search process for evolving improved collision-free robot move trajectories [13,14]. As we did not regard it as something special, we did not publish it in English (A detailed description in German of the cascaded weighted sum and all variants of the application of GLEAM to industrial robot path planning for several industrial robots can be found in [15].), until comments of reviewers of other publications on GLEAM changed our mind and revealed the need for a general discussion. Some impacts of the robot application on the evaluation of solutions are discussed later in Sections 4.2 and 4.3.3. Before the cascaded weighted sum is described, we will shortly introduce Evolutionary Algorithms and GLEAM in particular to the extent necessary for a better understanding of the following sections.

### 3.1. Short Introduction to Evolutionary Algorithms and GLEAM

Evolutionary Algorithms (EA) typically conduct a search in the feasible region, with this search being guided by a quality function usually called *fitness function*. The search is done in parallel by a set of solution candidates, called *individuals*, forming a *population*. If an individual is outside of the feasible region, it will be guided back by one or more penalty functions or comparable techniques. The fitness function may be one of the aggregation methods described above. Alternatively, it is guided by Pareto optimality and is therefore based on the amount of dominated solutions and possibly some other measure, which rewards a good spread of the non-dominated solutions along the Pareto front, see e.g., [10,12]. New solutions are generated by stochastic algorithmic counterparts of the two biological archetypes *mutation* and *recombination*, for which an individual selects a partner frequently influenced by the fitness. Thus, the generated offspring inherits properties from both parents. The third principle of evolution, the *survival of the fittest*, occurs when deciding about who is included in the next iteration. In elitist forms of EAs, the best individual survives always and therefore, the quality of the population can increase only. On the other hand, convergence cannot be ensured within limited time. To avoid premature convergence, various attempts have been made to maintain genotypic diversity for a longer period of time by establishing niches within the population, see e.g., [16–18]. GLEAM uses one of these methods [16] (An actual description of the diffusion model and its integration into GLEAM can be found in [19].) and, thus, frequently yields some distinct solutions of more or less comparable quality. Iterative, stochastic, and population-based optimization procedures in general tend to produce some variants of the best solution. How much they differ in properties and quality depends on the algorithm, the actions taken for maintaining diversity, and the length of the run.

### 3.2. Definition of the Cascaded Weighted Sum

In the *cascaded weighted sum* (CWS) each objective is assigned a weight $w_i$ as with the pure weighted sum and a priority starting with 1 as the highest one. If desired, some objectives may have the same priority. All objectives but those with the lowest priority receive a user-given threshold $\varepsilon_i$. In the beginning, only the objectives of the highest priority are *active* and contribute to the weighted sum. The others are activated according to the following priority rule:

> If all objectives with the same priority exceed their threshold,
> the objectives of the next lower priority are activated and their values are added to the sum.

As the objectives are grouped by the priorities and the groups are considered one after the other, the method is called *cascaded* weighted sum. A group whose members exceed their threshold is called a *satisfied group*. If at least one objective of a satisfied group drops below its threshold, the group is not satisfied anymore and consequently, all groups with lower priorities are deactivated, which will significantly reduce the resulting weighted sum.

For the formal definition of the CWS given in Equation (7), the original $f_i(x)$ are used for the threshold value checks rather than their normalized counterparts, as we assume that this is more convenient for experts of the application. The $k$ objectives are sorted according to their priorities and we have $g$ objective groups, where $1 < g \leq k$. For $g = 1$, the CWS would be identical with the weighted sum. Each group consists of $m_j$ objectives, the sum of which is $k$. As with the original weighted sum, each $w_i > 0$ and $\sum_{i=1}^{k} w_i = 1$.

As there are differences for the first and the last priority group, Equation (7) shows the objectives contributing to the weighted sum for the highest priority 1, the general case of priority $j$, and the lowest priority $g$.

**Priority 1**:     if not all $f_i(x) \geq \varepsilon_i \quad \forall i = 1, \ldots, m_1$
(highest priority)

$$\text{maximize } \sum_{i=1}^{m} w_i f_i^{\,norm}(x), \quad x \in S, \quad m = m_1$$

**Priority $j$**:     if all $f_i(x) \geq \varepsilon_i \quad \forall i = 1, \ldots, l_j$ and $\quad l_j = \sum_{l=1}^{j-1} m_l$   (satisfied groups)
$1 < j < g$

              not all $f_i(x) \geq \varepsilon_i \quad \forall i = l_j + 1, \ldots, l_j + m_j$             (7)

$$\text{maximize } \sum_{i=1}^{m} w_i f_i^{\,norm}(x), \quad x \in S, \quad m = l_j + m_j$$

**Priority $g$**:     if all $f_i(x) \geq \varepsilon_i \quad \forall i = 1, \ldots, l_g, \quad l_g = \sum_{l=1}^{g-1} m_l$         (satisfied groups)
(lowest priority)

$$\text{maximize } \sum_{i=1}^{k} w_i f_i^{\,norm}(x), \quad x \in S$$

Once a group is satisfied, the total quality value is increased abruptly by the values from the next activated group, which they can lose, if only one objective of a group with higher priority undergoes its $\varepsilon_i$. This makes it very unlikely for more successful solutions that the once gained values of the already contributing objectives drop below their thresholds in the course of further search.

The selection of appropriate weights and threshold values requires some knowledge about the problem at hand, including one or more preparative Pareto optimization runs as illustrated in the next section. Thus, neither the original weighted sum nor the CWS are a priori methods. We will come back to this later.

### 3.3. Example of the CWS

Table 1 gives an example of the usage of weights and thresholds for a problem of scheduling jobs organized in workflows of elementary operations to heterogeneous equipment comparable to the one described in [6]. All operations can be assigned to alternatively usable equipment at different costs and processing times. The task is to produce schedules, where the job processing is as cheap and as fast as possible and each job observes a given budget and a given due date. The rate of utilization of the equipment should be as high and the total makespan of all jobs as low as possible. Additionally, the schedules must be updated frequently, because e.g., new jobs arrive or waiting jobs are cancelled. As described in Section 2.2, all objectives are normalized according to Equation (3). The required limits are obtained as follows: The bounds of job time and costs are calculated by determination of the critical path of the workflow of that job and by the assignment of the fastest/slowest or costliest/cheapest equipment suited for the operations of a job. The user-given due dates and cost limits are checked against these bounds so that the subsequent scheduling is based on goals which are achievable in principle. The lower bound of the makespan is the duration of the longest lasting job using the fastest equipment and the upper bound is the sum of the duration of all jobs using the slowest equipment divided by the smallest number of alternatively usable equipment. As the rate of utilization already yields a value to be maximized between zero and one, there is no need for bounds.

**Table 1.** Example of the use of the cascaded weighted sum (CWS) and the effect of objective group weights. The objectives with the highest priority are always active and contribute to the weighted sum. They are marked here by a light green background.

| Priority | Objective | Weight [%] | Threshold $\varepsilon_i$ |
|:--------:|:---------:|:----------:|:-------------------------:|
| 1 | job time | 30 | 0.4 |
| 1 | job costs | 40 | 0.25 |
| 2 | makespan | 20 | - |
| 2 | utilization rate | 10 | - |

Job time and costs are most conflicting, while short processing times support a short makespan and tend to increase the utilization rate. Faster equipment typically is more expensive than slower, and the ratio between costs and duration of the use of equipment will play an important role. Thus, lower costs require the usage of equipment with a lower ratio of costs and duration. This tends to increase the duration and to decrease the workload of less cost-effective equipment. Additionally, shorter job times are also rewarded to some extent by the makespan and possibly by the utilization rate, but costs are not. These considerations are supported by the observation that the processing times are easier to reduce than costs. Thus, job time and costs should compete from the beginning and both should receive a larger portion of the weights. Therefore, they are grouped together at the highest priority so that they always contribute to the weighted sum, as shown in Table 1. As a rule of thumb, further objectives, which are less conflicting with each other and those of higher priorities, can go into the same group. After having determined the priority and grouping structure based on experience and considerations about the relationships between the objectives, appropriate weights and thresholds must be chosen.

Based on these considerations and a representative scheduling task, a schedule can be produced based on Pareto optimality for the identification of the region of interest, from which thresholds and weights can be derived. In the given example, this can be done in a first step for a reduced set of objectives by omitting a less conflicting one, e.g., the utilization rate. The Pareto fronts for good makespans of the two remaining objectives can be plotted. From it, the thresholds and the ratio of the weights between them can be derived, see Figure 2 for the relationship between Pareto front and the weights and Figure 5 for the usage of thresholds. This results in a ratio of 3:4 between the averaged job times and costs in the given example. The threshold values $\varepsilon_i$ are used as percentage values related to the available scale between $\min(f_i)$ and $\max(f_i)$, as shown in Equation (8):

$$f_{i,\varepsilon} = \min(f_i) + \varepsilon_i \left( \max(f_i) - \min(f_i) \right) \qquad (8)$$

In this case, the objectives of the next group are activated for schedules where the costs are below 75% of their available scales on the average and the finishing times are below 60% of their spendable time frames on the average. This approach can be repeated for the rest of the objectives or the remaining weights are assigned according to experience. For the given example, it was decided based on previous observations that about 70% of the weight should go to the first two objectives and the rest should go mainly to the makespan, as its reduction tends to increase the utilization rate. Table 1 shows the resulting weights. The suitability of the settings can be verified by the generation of a Pareto optimal schedule using all objectives and the comparison with the results obtained when using the CWS instead. Depending on the task at hand and the first setting of weights and thresholds, this may result in an iterative refinement.

To sum up, weights and thresholds are derived from experience and/or from previous estimations of the Pareto front of a representative task. In Section 4 we will discuss the range of meaningful applications of the CWS.

### 3.4. The Effect of the CWS on the Search

The effect of the cascaded assessment on population-based search procedures like EAs, PSO or ACO is illustrated in Figure 6 for two objectives and the example of the feasible objective region used in Figure 2. Based on previous knowledge, the sources of which are discussed in Section 4.3, a region of interest is defined for every objective group and the weights are set accordingly. Additionally, threshold values $\varepsilon_i$ are defined for all objectives but those of the group with the lowest priority. Care must be taken for the accessibility of the region of interest being not affected by these thresholds. In the example of Figure 6, objective two has a higher priority than objective one and a threshold value $\varepsilon_2$. In the beginning of the search, a quality gain can be achieved for upward moves only. For those solutions that have surpassed $\varepsilon_2$, the result of the first objective is added according to the weights changing the average movement direction towards the tangent and the region of interest. If the search runs long enough to come more or less close to convergence, most solutions will be found in the region of interest. The best of them will be at the intersection of the tangent and the Pareto front or at least close to it. Especially for EAs, which preserve genotypic diversity to some extent, good but suboptimal solutions close to the best one covering at least parts of the area of interest are very likely to be found. This means that a run is stopped when stagnation occurs over a longer period of time and not when the entire population has (nearly) converged.

**Figure 6.** Cascaded weighted sum for $k = 2$ and objective two having a higher priority than objective one. Thus, solutions in the hatched area are bettered according to $f_2$ only and will find the largest quality gain in upward moves (**red arrow**). This changes, if $\varepsilon_2$ is exceeded and $f_1$ starts to contribute to the resulting sum, as shown by the black arrows.

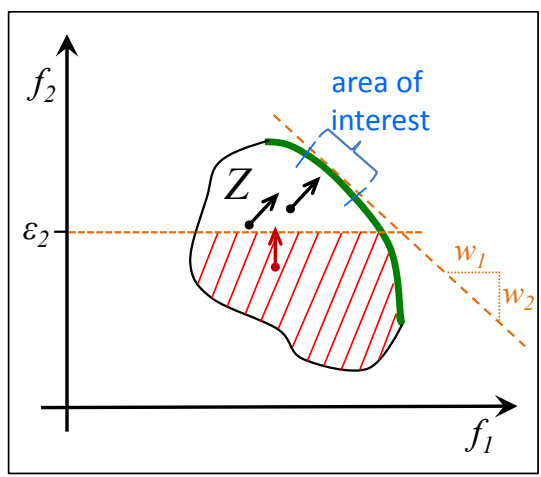

An example of a Pareto front with a non-convex section is shown in Figure 7 using the objective region and the threshold value of Figure 5. As the part between **F2** and the rightmost end of the Pareto front is quasi excluded, it is possible now to obtain solutions in the marked area of interest. This would not be the case for the original weighted sum. On the other hand, if the region of interest was located between the magenta dot and **F2**, most of the Pareto front would still be missed.

**Figure 7.** Cascaded weighted sum and region of interest for the example with a non-convex Pareto front given in Figure 5.

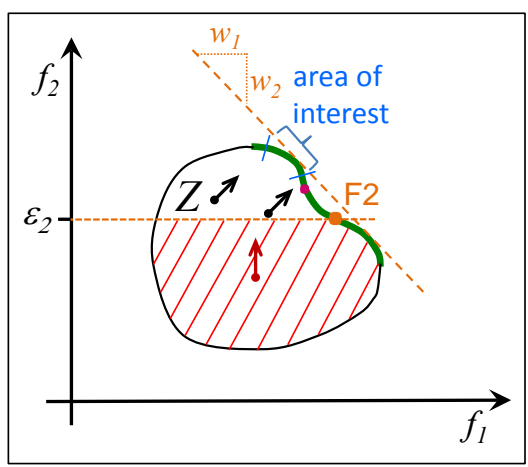

Normalization according to Equations (3) or (4) is done linearly with the same ratio for the entire interval $[\min(f_i), \max(f_i)]$. If previous knowledge is available for defining an area of interest for the Pareto front, corresponding subintervals are also known for the single objectives. This information can be used to tune the normalization function, as is shown exemplarily in Figure 8. More normalization functions can be found in [15].

**Figure 8.** Tuning the normalization of Equation (3) (**blue straight line**) to the interval of interest of one objective $f_i$. The decline outside of this interval is reduced drastically to allow for a strong increase inside, as shown by the green graph.

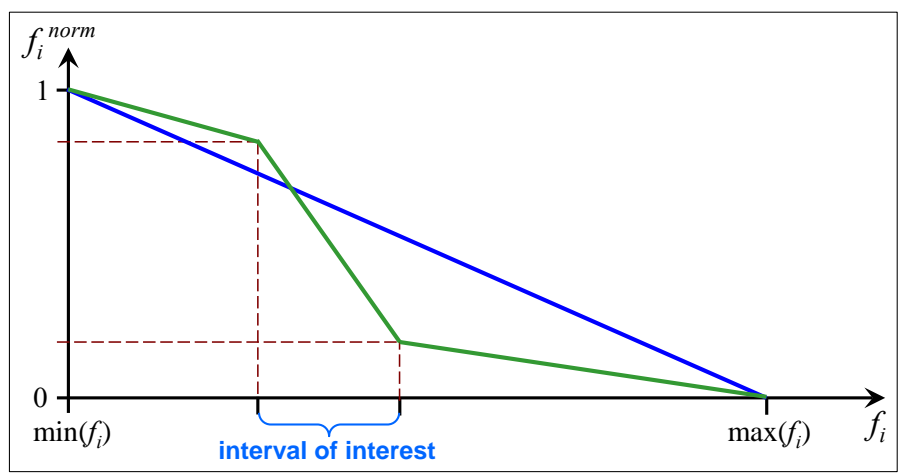

### 3.5. Summary

The grouping of the CWS reduces the amount of objectives considered at once and makes weighting easier. The CWS integrates objective thresholds comparable to those of the $\varepsilon$-constrained method, which are easier to handle for experts in the application field than weights, which now play a relatively minor role. The CWS allows for obtaining parts of a non-convex Pareto front, which were unreachable for the original weighted sum. However, it is still possible that some of these parts remain unattainable. These arguments underline the superiority of the CWS over the pure weighted sum. All aggregation methods alone are not suited as a priori approaches, as they require some previous knowledge to be parameterized.

## 4. Cascaded Weighted Sum and Its Field of Application

Optimization problems can be classified according to different criteria, such as the number of decision variables or of objectives, or the nature of the search space, where the number of (expected) suboptima or continuity plays an important role, or the type of project to which the optimization project belongs. The latter is often ignored in scientific literature, although it plays a significant role in real-world applications. Thus, we will take a closer look at that issue in the next sections. We will also consider the amount of objectives, as both properties are well suited to compare both assessment methods.

*4.1. Number of Objectives*

As already discussed in Section 3.3 objectives can conflict more or less. We consider here only objectives, the decision maker regards as conflicting in the sense that they shall be part of Pareto optimality. The amount of these objectives plays an important role for the practical applicability of the Pareto method. The Pareto front of up to three objectives can be visualized easily. For up to four or five objectives, decision maps, polyhedral approximation, or other visualization techniques can be used, see [20]. Interactive visualization techniques may support perception for more than three objectives, "but this requires more cognitive effort if the number of objectives increases", as Lotov and Miettinen summarize their chapter on visualization of the Pareto frontier in [20]. Thus, we can conclude that from five and in particular from more criteria, the perception and the comprehension of the Pareto front become increasingly difficult and turn into a business for experienced experts.

**Figure 9.** The number of required data points (Pareto-optimal solutions) of an approximation of a Pareto front increases exponentially with a growing number of conflicting objectives. The green line is based on a resolution of 7 data points per additional objective (axis), while the blue one uses 5 only.

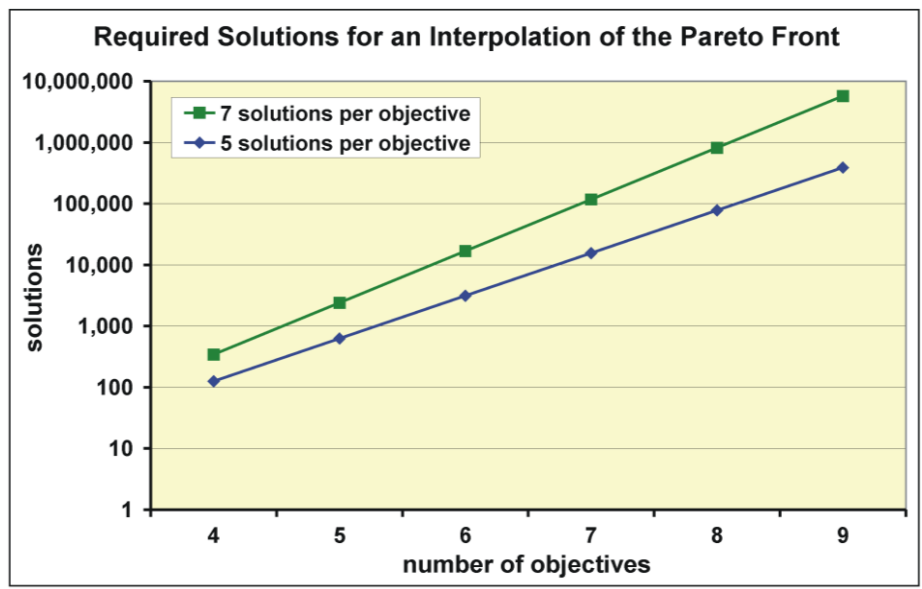

Another question is the effort to determine the Pareto front. For an acceptable visualization of the Pareto front, approximations like the one described in [21] may be used. Depending on the desired quality of approximation, a number of 5 to 7, in general $s$, Pareto-optimal solutions may be sufficient for two objectives. Assuming that the same quality of interpolation and granularity of support points shall be maintained when further objectives are added, $s^{(k-1)}$ support points are required for the interpolation of the hyperplane of the Pareto front, provided that all areas shall be examined. With interactive approaches, this can be reduced to some extent, but at the risk of missing promising regions. Figure 9 illustrates this growth of required Pareto optimal solutions. For 5 objectives, for example, 625 solutions are required to interpolate the entire hyperplane with 5 data points per axis. For a better interpolation quality obtained from 7 data points per axis 2401 solutions are needed. It should be noted that every data point requires several evaluations of solutions according to the optimization or approximation procedure used. Depending on the application, evaluations may be based on time-consuming simulations and last several seconds or minutes each. This clearly limits the practical applicability of the Pareto method for growing numbers of conflicting objectives. One solution is to reduce the number of objectives by an aggregation of less conflicting objectives to one.

*4.2. Classification of Application Scenarios and Examples*

Optimization projects can be classified into three different types:

I.   The nonrecurring type, which is performed once with little or no prior knowledge of e.g., the impact and relevance of decision variables or the behavior of objectives. This type requires many decisions regarding e.g., the number and ranges of decision variables, the number and kind of objectives, of restrictions, and more.

II.  The extended nonrecurring project, where some variants of the first optimization task are handled as well. Frequently, the modifications of the original project are motivated by the experience gained in the first optimization runs. As in the first type, decisions are usually made by humans.

III. The recurring type, usually based on experience gained from a predecessor project and frequently part of an automated process without or with minor human interaction only.

Examples of Types I and II are design optimization tasks like the design of micro-optical devices as described in [22] or problems from such a challenging field like aerodynamic design optimization, see e.g., [23]. A typical example of type III is the task of scheduling jobs to be processed on a computational grid as introduced in the last section and described in detail in [6]. Normally, nobody is interested in the details of the actual schedule, which usually will be replaced soon by a new one due to a replanning event like a new job to be planned or the introduction of new resources. Another example is the planning of collision-free paths for movements of industrial robots, as described in detail and for different industrial robot types in [14,15,24,25]. This example also shows that in some cases human judgment is possible in addition to a pure consideration of the achieved objective figures of a generated robot move path. The decision maker can take a look at the resulting movement using a robot simulator or the real device. Assessing this movement is much more impressive and illustrative than reading objective figures. On the other hand, such a well-fitting visualization is not always available.

*4.3. Comparison of Pareto Optimization and CWS in Different Application Scenarios*

4.3.1. Individual Optimization Project

For the first project type, the ranges of possibly achievable objective values usually are not known in advance. In this case, an estimation of them can be obtained from a Pareto optimization. From these data and the resulting Pareto front, a human decision maker can opt for additional and modified optimization or select the final solution. This type of optimization project clearly belongs to the domain of Pareto-based optimization.

### 4.3.2. Optimization Project with Some Task Variants

In many cases, the above statement also applies to the second project type, as experience is still limited and there must be good reasons to change the assessment method. Such a reason may be more than five objectives and if one or few areas of interest can be identified. In such cases, the computational effort can be reduced significantly. As mentioned before, an assessment of one solution in real world applications is frequently done by a simulation run, the duration of which strongly depends on the application at hand. One simulation may require seconds or even minutes and more. In such cases, the reduction of the number of evaluations is critical and an early concentration on the area of interest by using the CWS can be essential for the success of the project. On the other hand, the impact of the optimization is another important and application-dependent issue: If the savings expected from optimization justify the computational effort, Pareto optimization should be used until the areas of interest are reliably identified. Based on that, these areas can be explored in greater detail by optimization runs using the CWS, as illustrated in Figure 10. These considerations show that according to the project conditions, both methods may complement each other.

**Figure 10.** Both diagrams show a sample population of an advanced search more or less shortly before convergence. The CWS concentrates the best individuals (**black dots**) more or less on the region of interest, as shown in the left diagram. In contrast to that, Pareto-based optimization procedures attempt to distribute their solutions along the Pareto front as best as they can, see the right diagram. Thus, fewer solutions will be found in the area of interest.

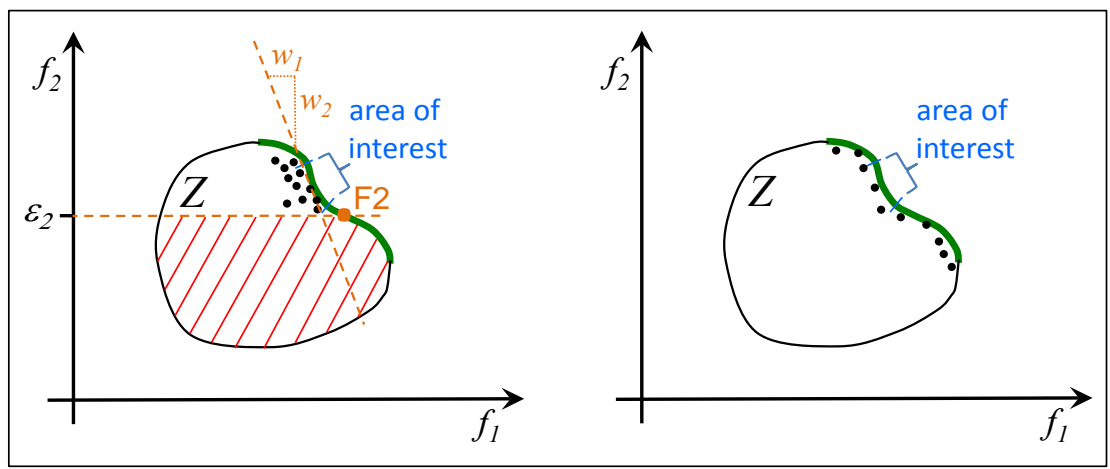

### 4.3.3. Repeated Optimization, also as Part of an Automated Process

Another domain of CWS-based optimization or planning is project type III, where the same task is executed repeatedly with minor modifications and, thus, known areas of interest. If a major change occurs, the area of interest can be adapted by using Pareto optimization. A typical example is the scheduling of jobs to the heterogeneous resources of a computational grid, as was introduced before and described in detail in [6]. It is a permanent replanning process, because events demanding a change of the schedule may occur long before a plan is completed. Examples are the introduction of new jobs or new resources, unexpected deactivations of resources, changes of the cost profiles of resources, early completion of jobs, or the like. As described in [6], five objectives are optimized and four penalty functions are used to handle the restrictions. Because planning time is limited and thousands of jobs and hundreds of resources must be handled, the planning must be stopped (long) before the used Evolutionary Algorithm converges. Thus, it is important to explore the region of interest as well as possible, see Figure 10. Additionally, there is no human expert to check the results several times per hour. For this automated scheduling process, the determination of the Pareto front makes no sense and the CWS is a meaningful alternative. These considerations also apply to many other scheduling tasks like the one described in [15,19].

Another example already mentioned is the planning of collision-free movement paths for an industrial robot [14,15,24,25]. Depending on the task at hand, we have four or five objectives and at the minimum one penalty function to handle collisions. As robot movements can be simulated and visualized, the results are checked by a human expert mostly on the level of robot movements rather than of objective figures. As areas of interest usually are also known in advance and new solutions should be generated fast, the CWS is suited here as well and for the same reasons as with the task before.

## 5. Conclusions

In Section 4.1 it was shown that the amount of solutions required to approximate a Pareto front increases exponentially with a growing number of conflicting objectives. As illustrated in Figure 9, the amount of evaluations increases considerably for more than five objectives. This limits the applicability of the Pareto approach for real-world applications, which frequently require time-consuming evaluations especially when based on simulation.

We have introduced the cascaded weighted sum (CWS), which can be described roughly as a combination of the weighted sum and the $\varepsilon$-constrained method. The major drawback of the pure weighted sum, the inaccessibility of parts of the Pareto front in non-convex cases, can be reduced to some extent by the CWS, see Section 3 and Figure 7. As the pure weighted sum, the CWS is not an *a priori* method. The major advantage of the CWS is its ability to concentrate solutions on the region of interest with less computational effort than Pareto optimization. In addition, this effort difference increases immensely with an increasing number of objectives, in particular for more than five. The region of interest can be a result of previous experience or knowledge, of a first Pareto-based optimization, or a combination thereof.

In Section 4.2, optimization projects were divided into three types, the individual project type, projects treating some variants of the task, and the type of repeated optimization of the same task with more or less small variations. The unquestioned domain of Pareto optimization is the first type of optimization projects. For the second type and five or more objectives, a combination of both methods can be advantageous as was described in Section 4.3.2. For the third project type of repeated optimization of task variants and with no or only minor human interaction, the Pareto front is not required, as the regions of interest are already known from previous solutions or initial Pareto optimization. The concentration of the CWS on that region is beneficial, as the computational effort can be reduced significantly. This is of importance especially in those cases, where a fast solution is required or the amount of evaluations is limited due to long run times.

Thus, we can conclude that both methods have their place and their field of application. Additionally, they can complement each other.

## Acknowledgments

We acknowledge support by the Deutsche Forschungsgemeinschaft and Open Access Publishing Fund of Karlsruhe Institute of Technology.

## Conflicts of Interest

The authors declare no conflict of interest.